\documentclass[journal]{IEEEtran}
\usepackage[utf8]{inputenc}
\usepackage{tikz}
\usepackage{tabularx, acro}
\usepackage{cite}
\usetikzlibrary{mindmap}
\usepackage[T1]{fontenc}
\usepackage{caption}
\usepackage{subcaption}

\DeclareAcronym{HSI}{ 
    short = {HSI}, 
    long  = {human-swarm interaction}
}

\DeclareAcronym{HRI}{ 
    short = {HRI}, 
    long  = {human-robot interaction}
}

\DeclareAcronym{UHRI}{ 
    short = {UHRI}, 
    long  = {underwater human-robot interaction}
}

\DeclareAcronym{UMRS}{ 
    short = {UMRS}, 
    long  = {underwater multi-robot system}
}

\DeclareAcronym{AUV}{ 
    short = {AUV}, 
    long  = {autonomous underwater vehicle}
}

\DeclareAcronym{UGR}{ 
    short = {UGR}, 
    long  = {underwater gesture recognition}
}

\DeclareAcronym{ML}{ 
    short = {ML}, 
    long  = {machine learning}
}

\DeclareAcronym{CV}{ 
    short = {CV}, 
    long  = {computer vision}
}

\DeclareAcronym{DL}{ 
    short = {DL}, 
    long  = {deep learning}
}

\DeclareAcronym{ROV}{ 
    short = {ROV}, 
    long  = {remote operated vehicle}
}

\DeclareAcronym{USBL}{ 
    short = {USBL}, 
    long  = {ultra-short baseline}
}

\DeclareAcronym{CADDY}{ 
    short = {CADDY}, 
    long  = {Cognitive Autonomous Diving Buddy}
}

\DeclareAcronym{VDD-C}{ 
    short = {VDD-C}, 
    long  = {Video Diver Dataset}
}

\DeclareAcronym{IMU}{ 
    short = {IMU}, 
    long  = {inertial measurement unit}
}

\DeclareAcronym{CNN}{ 
    short = {CNN}, 
    long  = {convolutional neural network}
}

\DeclareAcronym{R-CNN}{ 
    short = {R-CNN}, 
    long  = {region-based CNN}
}

\DeclareAcronym{EMG}{ 
    short = {EMG}, 
    long  = {electromyography}
}

\DeclareAcronym{LSTM}{ 
    short = {LSTM}, 
    long  = {long-short term memory}
}

\DeclareAcronym{SSD}{ 
    short = {SSD}, 
    long  = {single-shot detector}
}

\DeclareAcronym{HMM}{ 
    short = {HMM}, 
    long  = {Hidden Markov Models}
}

\DeclareAcronym{GMM}{ 
    short = {GMM}, 
    long  = {Gaussian Mixture Model}
}

\begin{document}
\title{Underwater Human-Robot and Human-Swarm Interaction: A Review and Perspective}

% \author{Sara~Aldhaheri
%         and~Giulia~De~Masi% <-this % stops a space
%         \\
%     Technology Innovation Institute, Abu Dhabi, United Arab Emirates\\
%     \{sara.aldhaheri, giulia.demasi\}@tii.ae \vspace{-1em}}

\author{Sara~Aldhaheri$^{\ast}$,  
 ~Federico~Renda$^{\dagger}$, and
        ~Giulia~De~Masi$^{\ast,\dagger}$
       % <-this % stops a space
        \\
    $^{\ast}$Technology Innovation Institute, Abu Dhabi, United Arab Emirates\\
    $^{\dagger}$Khalifa University, KU Center For Autonomous Robotic Systems, Abu Dhabi, United Arab Emirates\\
    \{sara.aldhaheri, giulia.demasi\}@tii.ae, federico.renda@ku.ac.ae \vspace{-1em}}
    
% The paper headers
% \markboth{Journal of \LaTeX\ Class Files,~Vol.~14, No.~8, August~2015}%
% {Shell \MakeLowercase{\textit{et al.}}: Bare Demo of IEEEtran.cls for IEEE Journals}

\maketitle

\begin{abstract}
% TODO The abstract goes here. 

There has been a growing interest in extending the capabilities of \acp{AUV} in subsea missions, particularly in integrating \ac{UHRI} for control. \Ac{UHRI} and its subfield, \ac{UGR}, play a significant role in enhancing diver-robot communication for marine research. This review explores the latest developments in \ac{UHRI} and examines its promising applications for multi-robot systems. With the developments in \ac{UGR}, opportunities are presented for underwater robots to work alongside human divers to increase their functionality. Human gestures creates a seamless and safe collaborative environment where divers and robots can interact more efficiently. By highlighting the state-of-the-art in this field, we can potentially encourage advancements in \ac{UMRS} blending the natural communication channels of human-robot interaction with the multi-faceted coordination capabilities of underwater swarms, thus enhancing robustness in complex aquatic environments.

%% Abstract archive which includes digital twins
% The emergence of numerous applications, datasets, and benchmarks within the domains of \ac{UHRI} and \ac{UGR} indicates a growing interest in research and innovation for marine applications. This review explores the latest developments in \ac{UHRI} and examines the potential role that digital twins for multi-robot systems may play in shaping the future of underwater swarms. With the developments in \ac{UGR}, opportunities are presented for underwater robots to work alongside human divers to increase their functionality. \Ac{UGR}, a topic of \ac{UHRI} focusing on improving the robot system's understanding of the diver, creates a seamless and safe collaborative environment where divers and robots can communicate more efficiently. By highlighting the state-of-the-art in this field, we can potentially pave the way for advancements in underwater multi-robot systems blending the natural communication channels of human-robot interaction with the multi-faceted coordination capabilities of underwater swarms, thus enhancing robustness in complex aquatic environments.

% \textit{Keywords - } Gesture recognition

\end{abstract}

\IEEEpeerreviewmaketitle

\section{Introduction}
\label{sec:0_introduction}
 
% This paragraph describes current uses for autonomous underwater robots and their advantages in replacing divers due to safety. In addition, there are advancements in technology that allow for extending their capabilities. % \IEEEPARstart{I}{n}

\Acp{AUV} are increasingly playing more crucial roles in assisting humans across different marine applications. Subsea missions such as data collection (e.g. \cite{wakita2010development,yuh2011applications}), ocean demining (e.g. \cite{teng2020underwater, mukherjee2011symbolic}), manipulation tasks (e.g. \cite{aldhaheri2022underwater,petillot2019underwater}), maritime infrastructure inspection (e.g. \cite{mcleod2013autonomous,segovia2020optimal}), and pipe maintenance (e.g. \cite{to2024underwater, zhang2021subsea, ho2020inspection, rumson2021application}) benefit from the added support of robots by monitoring surroundings, transporting hazardous objects, or performing other high-stress tasks.  % cite more from https://ieeexplore.ieee.org/stamp/stamp.jsp?tp=&arnumber=8962809
The need for autonomous robots proficient in these endeavors comes highly valued to human divers who are exposed to dangerous environments while experiencing high-cognitive load in handling these systems. Therefore, enabling human divers to effectively communicate with autonomous systems using gestures can enable intuitive interactions while  enhancing safety.

\begin{figure}[ht]
    \centering
    \includegraphics[width=0.5\textwidth]{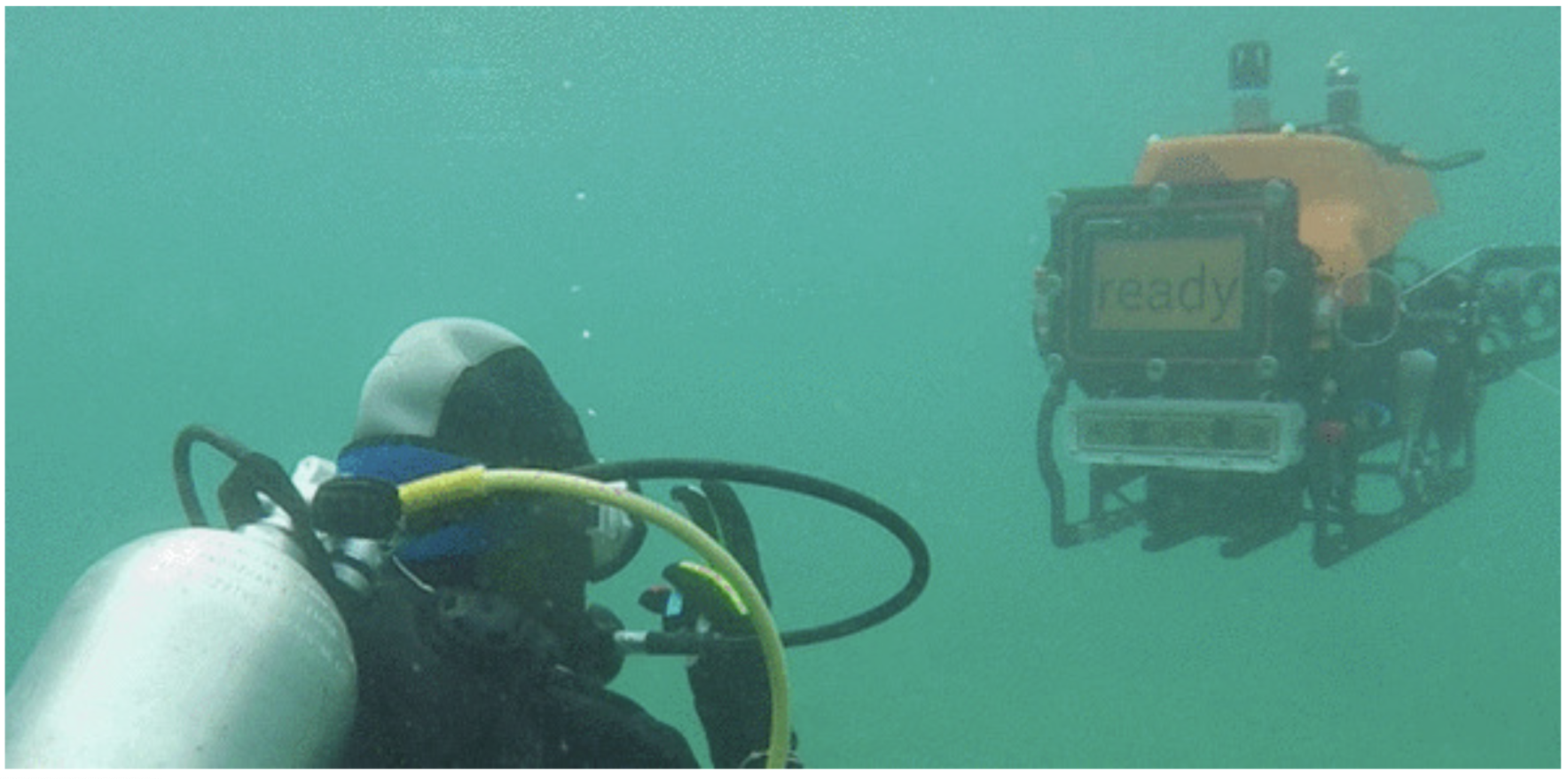}
    \caption{Diver using hand gestures to communicate command to underwater vehicle \cite{mivskovic2015caddy}.}
    \label{fig:uhri}
\end{figure}

In recent years, advancements in underwater robotic technology have not only broadened the horizons of oceanic explorations but have also ushered in a new era of seamless \acp{HRI}, more specifically \acp{UHRI} \cite{ mivskovic2016caddy, chavez2018robust, zhang2022underwater, sattar2018visual, islam2018dynamic,sattar2009underwater,chiarella2015gesture}. Central to this evolution is the integration of gesture recognition systems between human divers and underwater robots as an alternative means of machine control. Conventional remote controllers are cumbersome and would require  water-resistant setup with fast data transfer using joysticks \cite{birk2022survey}. Instead of relying solely on remote operations, divers and marine researchers can now communicate with robots using natural hand gestures (see Figure \ref{fig:uhri}) much like how they would communicate with fellow divers \cite{mivskovic2016human}. With such technology, we need to consider: \textit{how can we make \ac{UHRI} more reliable and how can we extend this capability to employ multi-robot systems}? As we transition from traditional, often complicated, remote controllers to more convenient gesture-based communication, analyzing this evolution becomes crucial for both current and future underwater endeavors.

% This paragraph should detail the limitations of current underwater HRI.
Current limitations of \acp{AUV} include their restriction to a singular function, such as object detection or tracking \cite{liu2022underwater}. This constraint arise from the reliance on traditional algorithms with low communication bandwidth and robustness asserting the need for a communication approach. Therefore, taking inspiration from the means of communication between divers, gesture recognition can be employed as a transformative solution, expanding the capabilities of \acp{AUV} beyond their current constraints. However, machine perception suffer from many factors such as refraction effects, scattering from turbidity, and color attenuation. Hence, although \ac{UHRI} promises an avenue for improving communication and control, there remain many challenges.

%------------------------------------------
\begin{figure*}
    \centering
    \includegraphics[width=0.9\textwidth]{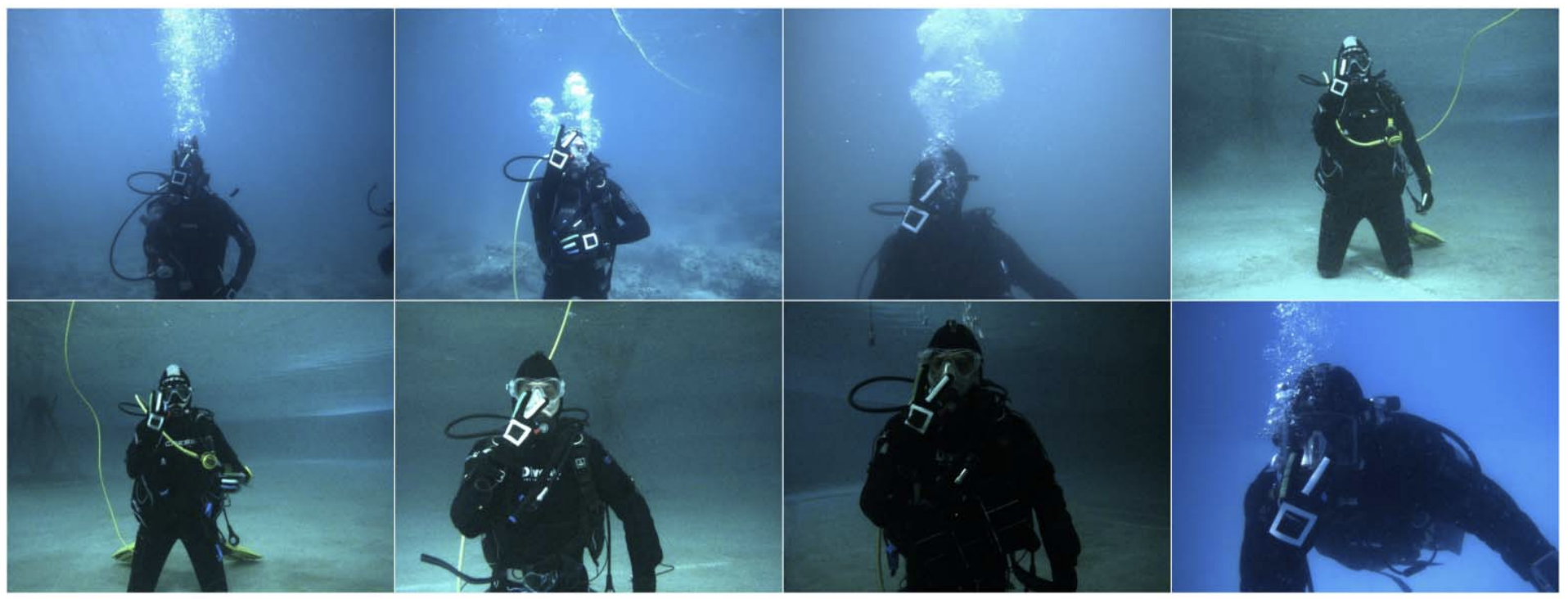}
    \caption{Hand gestures of diver used in the \ac{CADDY} dataset \cite{gomez2019caddy}.}
    \label{fig:gestures}
\end{figure*}
%------------------------------------------

% Conclude with a summary of the final paper content
In this paper, we present a comprehensive examination of the latest developments in \ac{UHRI} and its potential applications. Our focus encompasses a thorough exploration of human gesture recognition systems, delineating their capabilities and advancements in marine missions. Additionally, we provide an in-depth analysis of the diverse array of gesture recognition tools and scrutinize their limitations, underlying technologies, and prospects they offer in the domain of \ac{UMRS}, henceforth referred to as \textit{swarms}. Beyond the review, this manuscript discusses the integration of \ac{UHRI} for commanding swarms.

%%%%%%%%%%%%%%%%%%%%%%%%%%%%%%%%%%%%%%%%%%%%%%%%%%%%%%%%%%%%%%%%%%%%%%%%%%%%%%%%%%%%%%%%%%%%%%%%%%%%%%%%%%%%

% Introduce limitations of underwater swarm
% MOTIVATIONS behind using HRI underwater
%% despite having marine robots, many tasks remain dependent on human divers. 
%% MARKET DEMAND
%%% In addition to the market growth projections for global offshore repair and maintenance, the advancement of cutting-edge technologies for underwater environments further reinforces the motivations for human-robot collaboration \cite{kalyan2023concept}. %TODO \cite{https://www.fortunebusinessinsights.com/industry-reports/offshore-inspection-repair-maintenance-market-100405}
%% cutting-edge techs allowing for more extensions

% In this paper, we contribute a review of the vast potential applications and the state-of-the-art developments of \ac{UHRI}. We provide an extensive look into human gesture recognition systems and their offerings. Then, we detail the current use cases for understanding divers' gestures as well as the important topics of communication underwater. We will also elaborate on the diverse range of gesture recognition tools, encompassing both hardware and software solutions, and explore their potential limitations, underlying technologies, and the challenges and opportunities they present in the realm of \ac{UHRI}. Beyond the review, we provide clear directions for future research by addressing challenges hindering the development of \ac{UHRI}. Finally, this manuscript proposes a novel integration of \ac{UHRI} for instructing \ac{UMRS}. % TODO mention section

\section{A Review of Underwater Human-Machine Interaction}
\label{sec:1_review}
% provide summary of the use-cases, current applications and general interests in UHRI

This section explores the applications and advancements of \ac{UHRI} as well as its prospects in underwater swarm.  First, we identify and categorize the three most common algorithms handled by \ac{UHRI}. Then, we discuss examples of swarms that are currently handling \ac{HRI} in non-marine settings in order to understand how they can be integrated to the aquatic domain.

\subsection{Underwater Human-Robot Interaction}
In recent years, there has been a notable focus on advancing multimodal interactions between humans and \acp{AUV}, particularly through acoustic and optical sensing technologies \cite{ali2020recent,al2021underwater}. However,  the challenging aquatic medium exposes limitations in these modes such as the large signal attenuation for optical communication and low bandwidth for acoustic. Therefore, researchers have been exploring ways to enhance communication between human divers and underwater \acp{ROV} through visual inputs or motion changes from intuitive gesture commands. These underwater \acp{ROV} are generally employed to support various marine missions to complement the role of multiple divers. Hence, the literature reveals three main algorithms which hold potential for risk reduction using \ac{UHRI}: \textit{tracking}, \textit{monitoring}, and \textit{operations}. 

\textbf{Tracking.} The detection and tracking of one or multiple divers are fundamental roles of \acp{AUV} that hold avenues for \acp{UHRI}. Different methods of detection are implemented, namely acoustic or visual. Acoustic signals use localization methods involving surface vehicles and are useful for medium to long ranges \cite{demarco2013sonar,nadj2020using}. Furthermore, acoustic methods tend to use \ac{HMM} for clustering trajectories of moving blobs in the 2D sonar images. Visual target recognition and tracking uses frequency domain information \cite{sattar2007your} and are mainly operated in close ranges. With developments in computer vision and deep learning methods, tracking objects from monocular or stereo-cameras have become more prevalent. For example, algorithms like \acp{CNN}, \ac{R-CNN}, and Fast \ac{R-CNN} extract features from images to identify and classify between objects.  

\textbf{Monitoring.} Other algorithms implemented by \acp{AUV} are for sensing, mapping, and monitoring marine life, underwater structures, sea-floor, etc. High resolution 2D/3D mapping of submerged sites offer useful information to the area to be explored. Using photomosaicing techniques (e.g. visual odometry, SLAM, etc.) to chart visited areas allows robots to understand their environment \cite{casoli2021high, nocerino20213d}.

\textbf{Operations.} As stated in Section \ref{sec:0_introduction}, \acp{AUV} support underwater missions which involve handling a wide range of manual tasks. They can support in underwater welding, pipeline inspection, salvaging operations, and archaeological excavations \cite{khatib2016ocean, akkaynak2019sea}. By taking on these labor-intensive and often hazardous tasks, underwater robots enhance diver safety and efficiency. These robots are equipped with specialized tools and sensors, allowing them to execute precision tasks with greater accuracy and reliability. Subsea manipulation also include extraction of objects \cite{akkaynak2019sea}, demining \cite{teng2020underwater}, or valve handling \cite{aldhaheri2022underwater}.

By equipping \acp{AUV} with the capacity to comprehend and execute these three common commands issued by divers through conventional underwater communication channels, the initial challenges associated with diver safety, demanding tasks, and non-intuitive interaction will be effectively addressed. This integration of gesture recognition not only simplifies the interaction but also makes it more immediate and adaptable to dynamic underwater situations (see Figure \ref{fig:gestures}). From signaling a robot to collect a sample, maneuver around a coral reef, or follow a specific marine creature, gestures can provide instant, clear commands without the need for intermediary devices. Advanced sensors, coupled with sophisticated machine learning algorithms, enable these robots to discern a large range of gestures even in the challenging lighting and visibility conditions found underwater \cite{martija2020underwater}.
%%%%%%%%%%%%%%%%%%%%%%%%%%%%%%%%%%%%%%%%%%%%%%%%%%%%%%%%%%%%%%%%%%%%%%%%%%%%%%%%%%%%%%%%%%%%%%%%%%%%%%%%%%%%%%%%%%%%%%%%%%%%%%%%%%%%%%%%%%%%%%%%%%%%%%%%%%%%%%%%%%%%%%%%%%%%%%%%%%%%%%%%%%%%%%%%%%%%%%%%%%%%%%%%%%%%%%%%%%%%%%%%%%%%%%%%%%%%%%%%%%%%%%%%%%%%%%%%%%%%%%%%%%%%%%%%%%%%%%%%%%%%%%%%%%%%%%%%%%%%%%%%%%%%%%%%%%%%%%%%%%%%

\subsection{Human-Swarm Interaction} 
Land and aerial robot swarms presented their tremendous value to missions like search and rescue \cite{ruetten2020area,waharte2010supporting}, environmental monitoring \cite{carpentiero2017swarm,duarte2016application}, and surveillance \cite{orfanidis2019autonomous,abdelkader2021aerial}, where their ability to cover vast areas and possible dangerous unstructured environments is crucial. \Ac{HSI} is pivotal in leveraging these swarms, providing a means for humans to effectively manage and interact with large numbers of robots \cite{kolling2015human}. However, the challenge of autonomously monitoring dynamic targets in large, indeterminate environments means that relying solely on an autonomous swarm might not always yield the most effective results. Incorporating human intelligence into the process can markedly enhance a swarm's overall effectiveness. By integrating a human operator into the robot swarm system, there is a significant opportunity to leverage human cognitive abilities, such as strategic thinking and decision-making, thus substantially amplifying the swarm's capabilities. This human-robot synergy could lead to more efficient and effective outcomes, especially in complex scenarios where autonomous systems alone may fall short. 

%  Swarms have proven to be effective in exploring these areas with minimal human intervention in land and aerial settings.
% Metaverse and Digital Twins with underwater simulation for enhancing experiences
% multi-robot and metaverse (digital twins)
For underwater swarms, digital twins offer a remarkable advantage to safely understand the impacts of fusing \ac{HRI} to the system. They enable us to conduct tests and experiments in hazardous underwater settings without jeopardizing human divers. This technology enhances safety while optimizing robot behaviors and communication, ultimately advancing our capabilities for \ac{UHRI} \cite{kim2020user}.

\section{Underwater Human-Robot Interaction}
\label{sec:2_uhri}
% Include detailed descriptions of existing underwater human gesture dataset with limitations

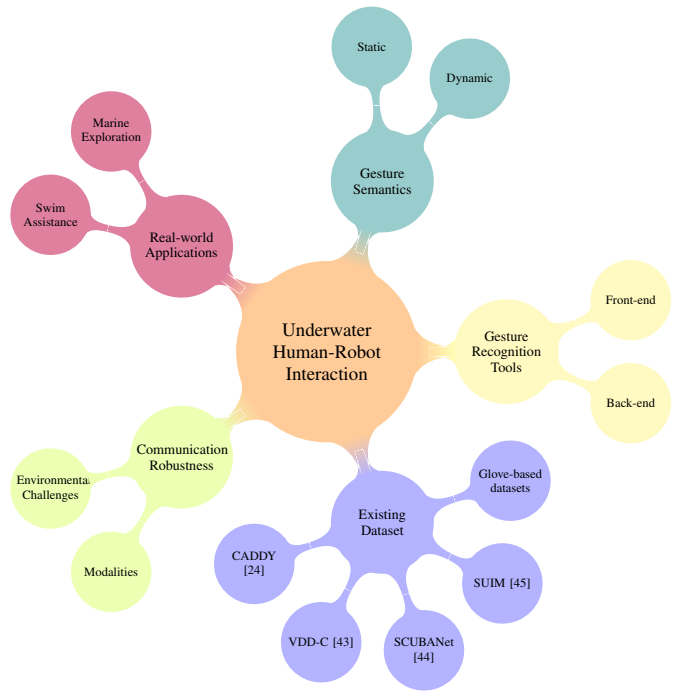
\begin{figure}[ht]
        \centering
        \resizebox{\columnwidth}{!}{
    \begin{tikzpicture}[mindmap, grow cyclic, every node/.style=concept, concept color=orange!40,
        level 1/.append style={level distance=4cm,sibling angle=72},
        level 2/.append style={level distance=3cm,sibling angle=45}]
    
    \node{Underwater Human-Robot Interaction}
        child [concept color=lime!30] { node {Communication Robustness}
            % child { node {Learning Models}}
            % child { node {Dataset Quality}}
            child { node {Environmental Challenges}}
            child { node {Modalities}}
        }
        child [concept color=blue!30] { node {Existing Dataset}
            child { node {CADDY \cite{gomez2019caddy}}}
            child { node {VDD-C \cite{de2021video}}}
            child { node {SCUBANet \cite{codd2019finding}}}
            child { node {SUIM \cite{islam2020semantic}}}
            child { node {Glove-based datasets}}
        }
        child [concept color=yellow!30] { node {Gesture Recognition Tools}
            child { node {Back-end}}
            child { node {Front-end}}
        }
        child [concept color=teal!40] { node {Gesture Semantics}
            child { node {Dynamic}}
            child { node {Static}}
        }
        child [concept color=purple!50] { node {Real-world Applications}
            child { node {Marine Exploration}}
            child { node {Swim Assistance}}
            % child { node {Underwater Equipment Maintenance}}
            % child { node {Data Collection}}        
            % child { node {Marine Exploration}}
            % child { node {Mapping and Surveillance}}
            % child { node {Underwater Equipment Maintenance}}
            % child { node {Data Collection}}            
        };
    \end{tikzpicture}}
        \caption{\Ac{UHRI} topics to consider when developing the technology. The literature covers the range of applications (red), the relevant diving gestures (teal), gesture recognition tools (yellow), existing dataset (purple), and robustness of this technology (green).}
        \label{fig:mindmap}
\end{figure}

This section discusses the several key facets related to \ac{UHRI} (see Figure \ref{fig:mindmap}). First, we categorize the range of gesture semantics, differentiating between static gestures, or a fixed pose, and dynamic gestures, or a sequence of movements. Building on this, an analysis of the diverse types of gesture learning tools is presented, highlighting the methodologies and technologies driving them. Furthermore, a critical examination of existing datasets is provided, emphasizing their composition, coverage, and relevance to underwater scenarios. Next, we evaluate the quality and robustness of both the computational models in use and the datasets they are trained on, ensuring a holistic understanding of the current state of underwater gesture recognition. Lastly,  we discuss the various applications and use-cases that exemplify the practical utility of this cutting edge technology in real-world maritime use-cases. 
% -------------------------------------------------------------------------------------------------------------

\subsection{Gesture Semantics}
With the growing work on \ac{UGR}, we recognize and present two types of gesture semantics: \textit{static} and \textit{dynamic}. Static gesture semantic refers to a non-moving pose that a diver presents to an assistant robot to convey a message. On the other hand, a dynamic gesture semantic refers pose that involves requiring motion tracking. 

\textbf{Static Gestures.} These gestures include steady poses by the diver either by one or both hands, or body poses. Commonly, a list of easily recognizable and natural hand gesture tokens are mapped into instruction tokens to improve the chances of discerning the intent of a diver \cite{islam2019understanding}. The pose must remain the same for a number of frames to differentiate static from dynamic gestures. The frames are then passed through a \ac{CNN} to segment regions for object detection accurately. 

% \begin{figure}[h!]
%     \centering
%     \includegraphics[width=\textwidth]{figures/diver-glove.png}
%     \caption{Diver's glove integrated with \ac{IMU}}
%     % \label{fig:enter-label}
% \end{figure}

\textbf{Dynamic Gestures.} 
Dynamic gestures are used to extend the lexicon for communicating with robots by including motion in the command classification. In this case, the change in pose is tracked to understand the movement. However, the challenge of ego-motion (motion of the camera) when identifying activities in the gestures of divers has to be addressed so as not to mistake these shifts with dynamic gestures. Work in \cite{buelow2011gesture} aims to resolve for the ego-motion through improved Fourier Mellin Invariant method (iFMI). The iFMI is a spectral registration method that recognizes the displacement of descriptors between consequent images.

% ------------------------------------------------------------------------------------------------------------- 

\subsection{Diver and Gesture Recognition Tools}
There are two main categories of \ac{UGR} tools associated with human-robot communication: \textit{front-end} and \textit{back-end}. Front-end algorithms are responsible for extracting features from diver gestures, whereas back-end algorithms handle the interpretation of the language used in these gestures. 

\textbf{Front-end.} Front-end tools typically involve machine perception methods that are well established in land, however, new models have to developed for underwater work. This is mainly due to the difference in the mediums. With developments in sensing and robotics, new learning architectures have emerged that allowed for progress in identifying divers and gestures underwater. Classical machine learning methods, like convex hull and support vector machines, are used to classify and recognize hand shapes \cite{chavez2021underwater}. Deep learning architectures that extract gesture include \ac{SSD} and Faster \acp{R-CNN} have achieved more than 90\% accuracy \cite{ren2015faster, chavez2021underwater}. Table \ref{table:dataset} highlights further approaches.

\textbf{Back-end.} Back-end tools, on the other hand, involve more intricate methods of mapping human intent with commands. For example, the work in \cite{yang2023dare}, an action recognition autoencoder is designed for robust and efficient underwater communication between divers and \acp{AUV}, addressing environmental, diver, and sensing uncertainties. It focuses on real-time recognition of diver actions to enhance safety and task completion in hazardous underwater conditions. The architecture processes stereo images of diver actions from an \ac{AUV}'s onboard camera, using deep bi-channel transfer learning to extract and fuse features for action identification. In \cite{agarwal2021predicting}, the \ac{VDD-C} dataset was used to develop a diver motion prediction model using \ac{LSTM} network. While in \cite{zhang2022underwater}, a text-visual model was developed to encode visual features from underwater diver images, textual features from category descriptions, and generate visual-textual features through multimodal interactions.

% -------------------------------------------------------------------------------------------------------------

%----------------------------------------TABLE-------------------------------------------
\begin{table*}[]
\caption{Comparison of \ac{UGR} datasets}
\resizebox{\textwidth}{!}{\begin{tabular}{lcllll}
\hline
\multicolumn{1}{c}{\textbf{Dataset}}                           & \textbf{Year} & \multicolumn{1}{c}{\textbf{\begin{tabular}[c]{@{}c@{}}Annotation \\ Type\end{tabular}}}                                           & \multicolumn{1}{c}{\textbf{Content}}                                                                                                                                              & \multicolumn{1}{c}{\textbf{\begin{tabular}[c]{@{}c@{}}Classification \\ Approaches\end{tabular}}}                                                                     & \textbf{Reference}                       \\ \hline
\Ac{CADDY}                                                     & 2014-2017     & \begin{tabular}[c]{@{}l@{}}Color markers on\\ standard diver-gloves\end{tabular}                                                  & \begin{tabular}[c]{@{}l@{}}12,708 stereo-images\\ of divers swimming du-\\ ring missions \\ 9,239 stereo-images \\ using gestures for com-\\ munication\end{tabular} & \begin{tabular}[c]{@{}l@{}}Convex hull,\\ Support Vector\\ Machine (SVM),\\ and combination\end{tabular}                                                              & \cite{abreu2015cooperative,birk2011co}   \\ \hline
SCUBANet                                                       & 2019          & \begin{tabular}[c]{@{}l@{}}Diver's body, head, \\ and hand both with \\ and without neoprene\\ glove of single color\end{tabular} & $\sim$1,000 stereo-images                                                                                                                                                         & \begin{tabular}[c]{@{}l@{}}Faster \ac{R-CNN} Inception\\ V2, Faster \ac{R-CNN} \\ ResNet101, \Ac{SSD} \\ MobileNet V2\end{tabular}                                                     & \cite{codd2019finding}                   \\ \hline        
\begin{tabular}[c]{@{}l@{}}Glove-based datasets\end{tabular} & 2019-Ongoing  & \begin{tabular}[c]{@{}l@{}}IMU,\\ Dielectric elastomer\\ (DE)\end{tabular}                                                         & \begin{tabular}[c]{@{}l@{}}12,480 samples for 13 \\ gestures (Antillon et al.)\end{tabular}                                                                                                         & \begin{tabular}[c]{@{}l@{}}Decision tree (DT), \\ SVM, Logistic Regres-\\ sor (LR), Gaussian \\ Naïve Bayes (GNB),\\ and Multilayer Perce-\\ ptron (MLP)\end{tabular} & \cite{antillon2022glove,walker2023diver} \\ \hline
SUIM                                                           & 2020          & \begin{tabular}[c]{@{}l@{}}Pixel level annotations\\ of various underwater\\ object classes\end{tabular}                          & \begin{tabular}[c]{@{}l@{}}$\sim$1,500 monocular \\ images of divers and other\\ various marine life\end{tabular}                                                                 & \begin{tabular}[c]{@{}l@{}}Presented SUIMNet - \\ fully convolutional \\ encoder-decoder sem-\\ antic segmentation of\\ salient objects\end{tabular}                  & \cite{islam2020semantic}                 \\ \hline
\Ac{VDD-C}                                                     & 2021          & \begin{tabular}[c]{@{}l@{}}Single and multiple divers\\ recorded in pools and open\\ water.\end{tabular}                          & \begin{tabular}[c]{@{}l@{}}$\sim$105,000 monocular \\ images\end{tabular}                                                                                                         & \begin{tabular}[c]{@{}l@{}}Faster R-CNN, \\ SSD with Mobilenet, \\ YOLO, and LSTM-SSD\end{tabular}                                                                    & \cite{de2021video}                      
\end{tabular}}
\label{table:dataset}
\end{table*}

\subsection{Existing Dataset and Limitations}
In this section, we delve into the cutting-edge underwater gesture recognition datasets (summarized in Table \ref{table:dataset}), exploring the latest advancements and methodologies employed in the field.

\subsubsection{CADDY}
Among the pioneering dataset to focus on diver-robot cooperation is the Cognitive Autonomous Diving Buddy, or CADDY \cite{abreu2015cooperative,birk2011co}. The CADDY dataset is a valuable resource aimed at advancing research in the domain of \ac{HRI} within underwater environments, particularly focusing on diver activities. This dataset is designed to facilitate the development and evaluation of robotic systems that can effectively collaborate with human divers in underwater scenarios. It comprises stereo-vision data collected from various underwater scenarios, capturing interactions between human divers and robots in diverse situations  (see Figure \ref{fig:gestures}). The stereo-images were captured using the Bumble-
Bee XB3 stereo RGB camera system  (resolution of 640×480 pixels) \cite{antillon2022glove}. The dataset provides detailed information on the underwater environment, including the presence of divers, their activities, and the interactions with robots, all of which are crucial for training and testing \ac{HRI} algorithms and systems. Researchers utilize the CADDY dataset to enhance the capabilities of underwater robotics systems, improve safety during diver activities, and explore innovative applications in underwater exploration and intervention \cite{chavez2021underwater}.

\subsubsection{SCUBANet}
The SCUBANet dataset consists of images of object classes commonly encountered during such interactions, with a specific focus on diver detection. This dataset consists of approximately 2000 labelled images of divers, hands, and heads. The images in SCUBANet are annotated with per-instance bounding boxes, providing a valuable resource for developing and enhancing diver detection algorithms. The dataset was created using crowd-sourced annotations through a web-based interface, aiming to facilitate deployment in real-world underwater robotics applications\cite{codd2019finding}. 

\subsubsection{Glove-based Datasets}
The underwater environment exerts various influences on underwater images, resulting in challenges such as noise interference, the refraction effect, wavelength color attenuation, and poor visibility. Consequently, accurately identifying diver images captured by \acp{AUV} in intricate underwater conditions becomes difficult. An alternative approach involves employing \ac{EMG} sensors which record muscle responses or electrical activity elicited by nerve stimulation of the muscle. These sensors are placed on the skin's surface to detect electrical signals in the arm, classifying them into hand movements. Specific hand movements are mapped to particular instructions which are understood by the robot system. Nevertheless, implementing this method underwater poses difficulties, primarily because the EMG signals can weaken, and finding the right sensor placement can be challenging. Therefore, efforts towards developing gloves integrated with \acp{IMU} have been undertaken across different projects \cite{antillon2022glove,tang2023diving}.

\subsubsection{SUIM}
The SUIM dataset is the first large-scale dataset for semantic segmentation of underwater imagery and contains over 1500 images with pixel-level annotations for eight object categories, including fish, reefs, aquatic plants, wrecks/ruins, human divers, robots, and the sea-floor. These images, collected during oceanic explorations and human-robot cooperative experiments, are essential for developing and testing semantic segmentation models in underwater environments. The SUIM dataset addresses issues like unique image distortion artifacts and the lack of large-scale annotations by providing a comprehensive set of annotated images for a variety of underwater object categories. Additionally, SUIM gave rise to the SUIM-Net model, a fully convolutional encoder-decoder architecture with skip connections, designed to effectively perform semantic segmentation on the SUIM dataset\cite{islam2020semantic}.

\subsubsection{\ac{VDD-C}}
The \ac{VDD-C} is a significant dataset containing approximately 105,000 annotated images of divers underwater, compiled in 2021 by researchers at the University of Minnesota. This dataset, collected from videos of divers in pools and off the coast of Barbados, is designed to advance diver detection algorithms for AUVs. The sequential nature of the images, derived from videos, makes them ideal for training temporally aware algorithms. The dataset, which expanded the training set size by 17 times, has significantly improved existing diver detection algorithms \cite{de2021video}. This dataset facilitates the training and evaluation of various state-of-the-art object detection networks, such as \ac{SSD} with MobileNet, Faster \ac{R-CNN}, and YOLO, including their video stream variants \cite{jesus2022underwater}. The paper evaluates these networks not only based on traditional accuracy and efficiency metrics but also examines their temporal stability in detections, a critical factor in dynamic underwater environments. Additionally, the study delves into analyzing the failure modes of these detection systems, providing insights into their limitations and guiding future improvements in diver detection technology for \ac{AUV} applications.

% %------------------------------------------
% \begin{figure*}
%     \centering
%     \includegraphics[width=0.75\textwidth]{figures/diverglove.png}
%     \caption{IMU-integrated gloves for gesture recognition}
%     \label{fig:gloves}
% \end{figure*}
% %------------------------------------------

% -------------------------------------------------------------------------------------------------------------

\subsection{Communication Robustness}
% Describe the difficulty of using USBL and why it's actually better to use HRI. Discuss accuracies of acoustic and inconvenience of tethered systems
% Acoustic data transmission for communication and navigation remains an essential requirement to ensure the security of human-robot interactions. 
Underwater communication robustness is a burgeoning field that demands further development to address the unique challenges posed by underwater environments and enhance the reliability of communication   systems. Among these challenges are environment and modality.

\textbf{Environment. } Many factors affect perception underwater which may lead to unreliable communication especially with a system relying on vision for relaying information. Depth, turbidity, and salinity heavily impact the performance of models. For example, minimal light penetrates past 200 meters and significantly drops beyond 1000 meters. Furthermore, highly turbid waters increase light attenuation and scattering intensity rendering optical systems ineffective \cite{huy2023object}. Salinity has a similar effect as saltier water causes a bigger electromagnetic wave attuenuation. 

\textbf{Modality. } One of the primary obstacles revolves around inter-unit communication, as acoustic communication grapples with limitations like restricted bandwidth, latency, and sporadic interruptions inherent in the propagation of acoustic waves \cite{fulton2019robot}. In addition, the use of \ac{USBL} systems also presents a  challenge in underwater operations, as they rely on acoustic signals and can be susceptible to interference and inaccuracies in complex environments. Tethered systems, while offering communication stability, introduce the inconvenience of physical constraints and limitations on mobility for both humans and robots, making the development of more intuitive and flexible \ac{HRI} solutions all the more imperative. 

\subsection{Real-world Applications}
As discussed in Section \ref{sec:0_introduction}, \ac{UHRI} allows diver to convey commands to \acp{AUV} using gestures as is the case with human-human underwater communication \cite{council2014common}. However, the real-world application of \ac{UHRI} remains quite rare. Regulating underwater robots based on human inputs and facilitating tool hand-off with a human diver \cite{demarco2014underwater} are examples of tasks that diver can employ \ac{UHRI}. These capabilities can serve in underwater equipment maintenance. Additionally, divers are able to instruct \acp{AUV} in mapping and surveying seafloors or submerged artifacts \cite{islam2020semantic}. Another illustration of \ac{UHRI}'s practicality lies in data collection. Researchers have developed a sophisticated framework that combines human input and existing environmental knowledge to plan efficient routes for gathering scientific data in marine environments \cite{somers2016human}. Furthermore, another study examines methods for robots to effectively communicate their findings and operational status to human divers during missions \cite{fulton2019robot}.

% -------------------------------------------------------------------------------------------------------------

%Therefore, when considering communication robustness, there are a number of challenges to observe. These challenges can be \textit{environmental} or \textit{algorithm/system-related}.  

% Figure \ref{fig:caddyconcept} depicts the multiple components that comprise the CADDY human-robot interface: the surface vehicle, the underwater robotic assistant and the diver.

% \begin{figure}[h!]
%     \centering
%     \includegraphics[width=0.35\textwidth]{figures/concept.png}
%     \caption{The CADDY multicomponent concept comprising a human diver, a surface vehicle and an \ac{AUV} \cite{mivskovic2015overview}.}
%     \label{fig:caddyconcept}
% \end{figure}

\section{\ac{UHRI} for Swarms}
\label{sec:3_multi}

\begin{figure}
     \centering
     \begin{subfigure}[b]{0.3\textwidth}
         \centering
         \includegraphics[width=\textwidth]{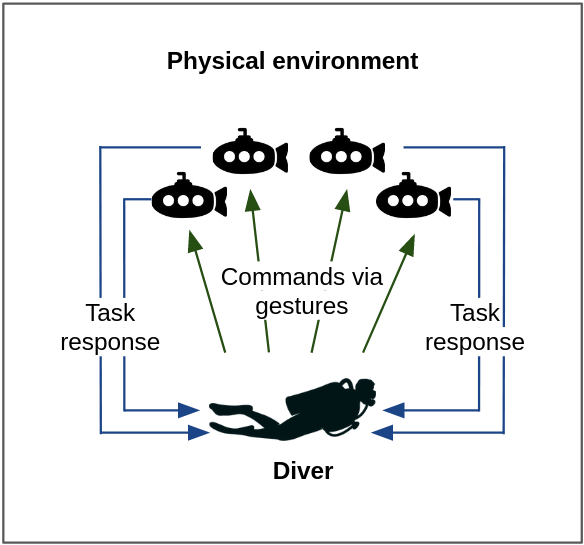}
         \caption{Individual robot control}
         \label{fig:individual}
     \end{subfigure}
     \hfill
     \begin{subfigure}[b]{0.3\textwidth}
         \centering
         \includegraphics[width=\textwidth]{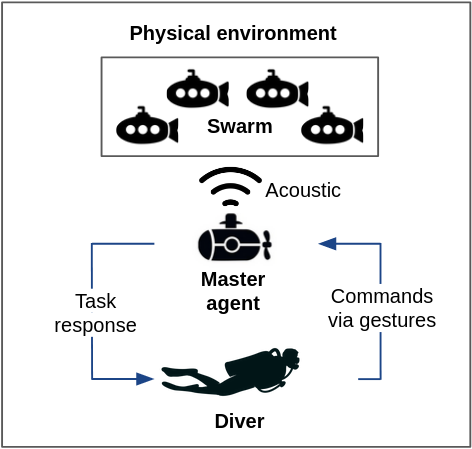}
         \caption{Hierarchical swarm control}
         \label{fig:hierarchical}
     \end{subfigure}
     \hfill
        \caption{Individual control of robots in (a) compared to multi-level autonomy in (b).}
        \label{fig:proposedframework}
\end{figure}

The study of underwater swarms has seen increasing prevalence in recent years \cite{luvisutto2022robotic, Connor21}. While considerable advancements have been made in underwater robot technology, the vastness of the ocean presents challenges for a single robot's scope mainly due to limitations in their operational power. Consequently, there is a growing desire among researchers to deploy underwater swarms to achieve tasks with greater efficiency and cost-effectiveness \cite{iacoponi2022h}. One approach towards this objective is to incorporate a multi-level, or hierarchical, framework where communication will only be required to be made with one master agent as opposed to each \ac{AUV} individually. Such architecture comprises a master agent that conveys task commands to multiple swarm agents expanding the work in \cite{kolling2015human} to the marine domain (see Figure \ref{fig:proposedframework}). This work draws inspiration from the concept of \textit{shepherding}, a method of communicating with multiple agents to accomplish missions demanding more than from a single robot system. Extending on this approach, communicating with the master agent using \ac{UGR} will endow swarms with the aforementioned advantages of \ac{UHRI}. In essence, the combination of \ac{UHRI} with swarm introduces a novel method of \ac{HSI} for controlling multiple agents underwater.

A more novel and visionary approach for enhanced supervision of marine swarm missions can be through simulation. The integration of sensors in \acp{AUV} to feed in data to digital twins signifies an innovative shift in underwater human interaction within the context of divers and swarms \cite{cichon2018digital}. Digital twins, which refer to virtual representations of physical entities, offer a unique perspective in enhancing the interaction between divers and robots in the underwater environment. Through real-time simulation and monitoring, digital twins enable a more comprehensive understanding of the underwater work space, providing valuable insights for both human divers and robotic systems \cite{lambertini2022underwater, tao2018digital}.

\begin{figure}
    \centering
    \includegraphics[width=0.5\textwidth]{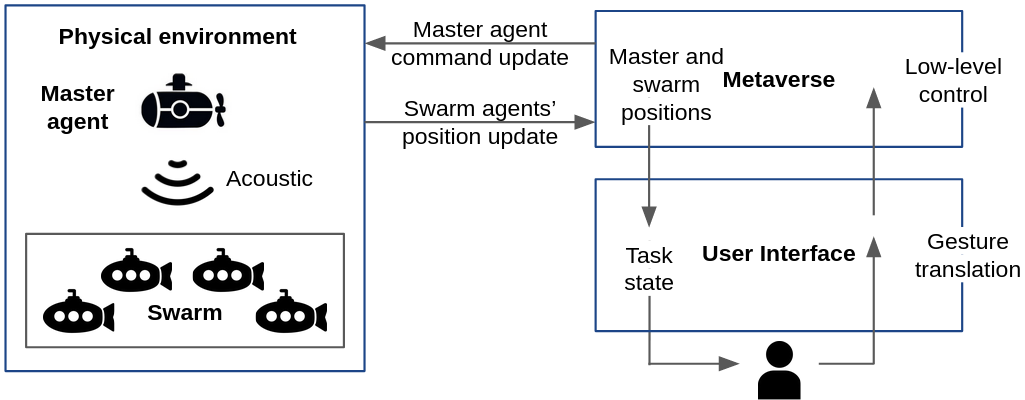}
    \caption{Suggested metaverse framework using \ac{UHRI}.}
    \label{fig:metaverse}
\end{figure}

In parallel, the concept of a metaverse, characterized by a collective virtual shared space, introduces new dimensions to human-robot collaboration \cite{nguyen2023swarm}. As human divers engage in immersive experiences within the metaverse, they can seamlessly interact with and control swarms deployed in the underwater domain. This immersive and interconnected environment not only enhances the divers' situational awareness but also facilitates more intuitive and responsive communication with robotic counterparts. This approach paves the way for digital twin-enabled metaverse, simulation combining real and virtual worlds, enhancing scalability in \ac{HSI} (see Figure \ref{fig:metaverse}). Such framework comprise a user interface through which a command can be gestured. The instruction is then translated into a low-level command and the master agent is updated. The master agent collects the location data from each individual agent and relays this information to the metaverse, which facilitates human oversight. Incorporating a human into the feedback loop is considered essential to accommodate human-like intelligence and is a requisite for ensuring ethical and safe \ac{HSI}.  The efficacy of \ac{HSI} relies on the presence of a control agent that is superior in both physical and cognitive aspects. Nevertheless, it is crucial to take into account the disparity in simulation quality between land-based and marine environments, stemming from communication delays, lack of active maintenance and the limited availability of realistic simulations in the latter \cite{alvarez2019generation}.

%%%%%%%%% swarm metaverse for multi-level autonomy using digital twins       
% The concept of a 'swarm metaverse' is a visionary approach to managing and interacting with robot swarms. This idea leverages a virtual environment, or 'metaverse', where digital twins of real-world robotic entities exist \cite{nguyen2023swarm}. These digital twins mirror the physical characteristics and behaviors of their real-world counterparts, providing a unique and efficient platform for human operators to engage with and control the swarm. Recent work towards combining the two ideas has materialized in \cite{nguyen2023swarm}.

\section{Discussion and Perspective}

In summary, we categorized the functions of \ac{UHRI} to include tracking, monitoring, and handling tasks in Section \ref{sec:1_review}. Section \ref{sec:2_uhri} elaborates on the real-world applications of \ac{UHRI}, types of gesture semantics used for \ac{UGR} in addition to the state-of-the-art tools and dataset implemented for secure human-diver communications. The integration of gesture recognition in underwater robots signifies more than just a technological achievement; it represents a paradigm shift in how humans and machines collaborate in the marine environment \cite{chavez2021underwater, chiarella2015gesture, liu2022underwater, yang2019diver}. By making this interaction as natural and seamless as possible, we not only enhance the efficiency of underwater tasks but also improve reliability between divers and their robotic counterparts. As we navigate this promising frontier, continued research and innovation in \ac{UGR} will undoubtedly play a pivotal role in shaping the future of underwater exploration and collaboration. The discussed integration of metaverses with \ac{UHRI} in Section \ref{sec:3_multi} represents an exciting avenue for advancing underwater robotics. Digital twins, the immersive real-time feed of an environment space, can serve as powerful simulation platforms for training, testing, and optimizing the performance of underwater robots. Furthermore, these simulations can facilitate remote collaboration for divers. Expanding on this notion, the concept of a 'swarm metaverse' is a visionary approach to managing and interacting with robot swarms. This idea leverages a virtual environment, or 'metaverse', where digital twins of real-world robotic entities exist. These digital twins mirror the physical characteristics and behaviors of their real-world counterparts, providing a unique and efficient platform for human operators to engage with and control the swarm. Recent work towards combining the two ideas has materialized in \cite{nguyen2023swarm}.

With regard to controlling these swarms, there are a number of mechanisms varying in degrees of autonomy employed in ground and aerial settings that can be translated to marine applications. For instance, teleoperating swarm members offer simple robot control but suffers from a disproportionately high human-to-robot ratio and a consequential exponential increase in the workload required for operation \cite{velagapudi2008scaling, crandall2005validating}. This limitation serves as a motivation for using a hierarchical control method as explained in Section \ref{sec:3_multi}. On the other hand, parameter selection allows for adaptable swarm behaviors, yet its application is constrained by its incompatibility with real-time or online deployment. Behavior selection, while user-friendly for novices, necessitates preprogramming of all behaviors and lacks the flexibility to respond to unforeseen events, with a particular sensitivity to the timing of behavior switching. Hence, having a supervising human input would improve task performances. Finally, controlling a few swarm members allows for significant human intervention, hence, it is the most reliable \ac{HSI} mechanism. Nevertheless, it is susceptible to adversarial attacks targeting swarm leaders and exhibits difficulties in handling diverse swarm configurations.

%%%%%%%%%%%%%%%%%%%%%%%%%%%%%%%%%%%%%%%%%%%%%%%%%%%%%%%%%%%%%%%%%%%%%%%%%%%%%%%%%%%%%%%%%%

\section{Final Remarks}

This \ac{UHRI} and underwater \ac{HSI} review encapsulates the rich avenues for further research of this dynamic field. By delving into key facets such as the gesture semantics, communication modalities, hierarchical control, and the challenges of these topics, we recognize the current extents and limitations of \ac{UHRI}. As underwater research progresses, new findings will continue to evolve and bridge current gaps between marine technology and their more established ground and aerial counterparts particularly in swarm robotics.

% This manuscript has provided a comprehensive overview of the state-of-the-art applications of \ac{UHRI} categorizing the various use-cases and gesture semantics found in the literature. This paper also highlights the prevalent datasets, encompassing both visual and motion inputs. The discussion further delves into the crucial concepts surrounding communication robustness and data handling tools for these datasets.
% Additionally, we explore an extension of swarm technology into the underwater domain, emphasizing the potential integration of gesture-based interaction, a well-established technique in terrestrial and aerial environments. This proposal envisions the utilization of digital swarm technology and metaverses to streamline system monitoring, offering a promising avenue for future research and practical implementation in the underwater realm.

\section*{Acknowledgments}
The authors acknowledge the project “Heterogeneous Swarm of Underwater Autonomous Vehicles”, a collaborative research project  between the Technology Innovation Institute (Abu Dhabi) and Khalifa University  (contract no. TII/ARRC/2047/2020).

\bibliographystyle{IEEEtran}
\bibliography{bibtex/bib/references}

\end{document}